\documentclass{article}
\usepackage{amsmath,graphicx,mlspconf}
\usepackage{cite}
\graphicspath{./images}
\usepackage{subfigure}
\usepackage[font={scriptsize}]{caption}
\copyrightnotice{U.S.\ Government work not protected by U.S.\ copyright}

\copyrightnotice{978-1-7281-6662-9/20/\$31.00 {\copyright}2020 Crown}

\copyrightnotice{978-1-7281-6662-9/20/\$31.00 {\copyright}2020 European Union}

\copyrightnotice{978-1-7281-6662-9/20/\$31.00 {\copyright}2020 IEEE}

\toappear{2020 IEEE International Workshop on Machine Learning for Signal Processing, Sept.\ 21--24, 2020, Espoo, Finland}

\title{Low-Light Environment Neural Surveillance}
\name{Michael~Potter, Henry~Gridley, Noah~Lichtenstein,
        ~Kevin~Hines, John~Nguyen, and
        Jacob~Walsh\thanks{Thanks to advisor Northeastern University Professor Bahram Shafai as our advisor.}}
\address{Northeastern University \\ Electrical and Computer Engineering \\ \{potter.mi,gridley.h,lichtenstein.n,hines.ke,nguyen.john,walsh.jac\}@northeastern.edu}

\begin{document}

\maketitle

\begin{abstract}
We design and implement an end-to-end system for real-time crime detection in low-light environments. Unlike Closed-Circuit Television, which performs reactively, the Low-Light Environment Neural Surveillance provides real time crime alerts. The system uses a low-light video feed processed in real-time by an optical-flow network, spatial and temporal networks, and a Support Vector Machine to identify shootings, assaults, and thefts. We create a low-light action-recognition dataset, LENS-4, which will be publicly available. An IoT infrastructure set up via Amazon Web Services interprets messages from the local board hosting the camera for action recognition and parses the results in the cloud to relay messages. The system achieves 71.5\% accuracy at 20 FPS. The user interface is a mobile app which allows local authorities to receive notifications and to view a video of the crime scene. Citizens have a public app which enables law enforcement to push crime alerts based on user proximity.
\end{abstract}
\begin{keywords}
Computer vision, Low-light, Crime detection, Amazon Web Services
\end{keywords}

\section{Introduction}
Intelligent surveillance cameras have become increasingly popular in both commercial and private applications. Traditionally, Closed-Circuit Television (CCTV) systems are the de facto monitoring system. More recent products such as Ring home security systems use Internet of Things (IoT) connected surveillance video feeds to provide homeowners an extra degree of security through remote monitoring. Most existing monitoring systems require some amount of manual operation and have limited use in low-light environments. Other similar solutions include the Code Blue emergency call system, which many universities have deployed on their campuses. This system provides a direct communication line to first responders. However, the call station depends on the user being able to reach and physically push the call button. Furthermore, the rate of reported crimes is dependent on the victims or bystanders to self-report. 

Though there exist algorithms for fully automated action recognition \cite{DBLP:journals/corr/two_stream,DBLP:journals/corr/temporal_networks,DBLP:journals/corr/spatio_temporal_conv,DBLP:journals/corr/real_time_action_rec,DBLP:journals/corr/quo_vadis}, many are not applied in real-time or low-light environments. The existing benchmark action recognition datasets such as HMDB-51 \cite{Kuehne:2011:HLV:2355573.2356424} (Human Motion DataBase), UCF-101  \cite{DBLP:journals/corr/abs-1212-0402} (University of Central Florida), and Sports-1M \cite{karpathy2014large} contain primarily daytime videos. UCF released the UCF-Crime dataset \cite{DBLP:journals/corr/abs-1801-04264} for general anomaly detection and recognizes 13 crime categories, including \textit{arrest, arson, assault, burglary, explosion, fighting, robbery, shooting, stealing, shoplifting, and vandalism}. This dataset consists of indoor video feeds or camera views focused on buildings, and untrimmed videos with  noisy labels which makes the dataset hardly viable for real-life application. 

    \begin{figure}[!h]
        \centering
        \includegraphics[width=7cm]{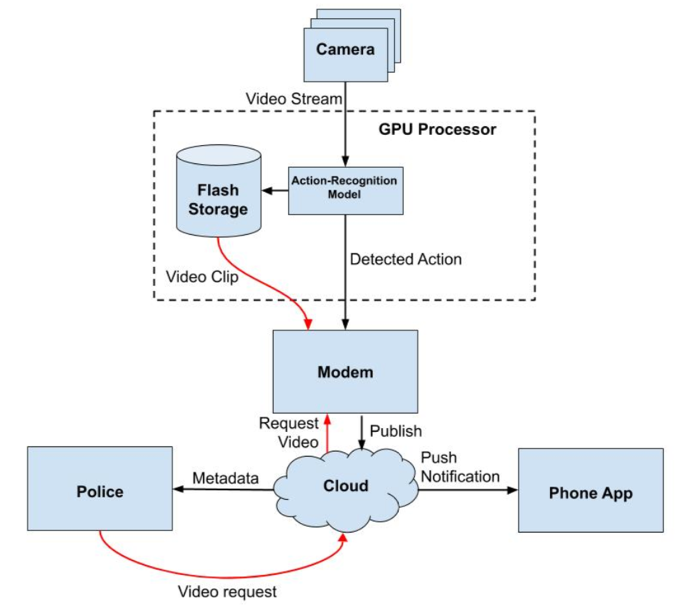}
        \caption{LENS system architecture diagram}
        \label{fig:system_diagram}
        \vspace{-.5em}
    \end{figure}

\begin{figure*}[ht]
\begin{minipage}[b]{0.7\linewidth}
    \centering
    \includegraphics[width=12cm]{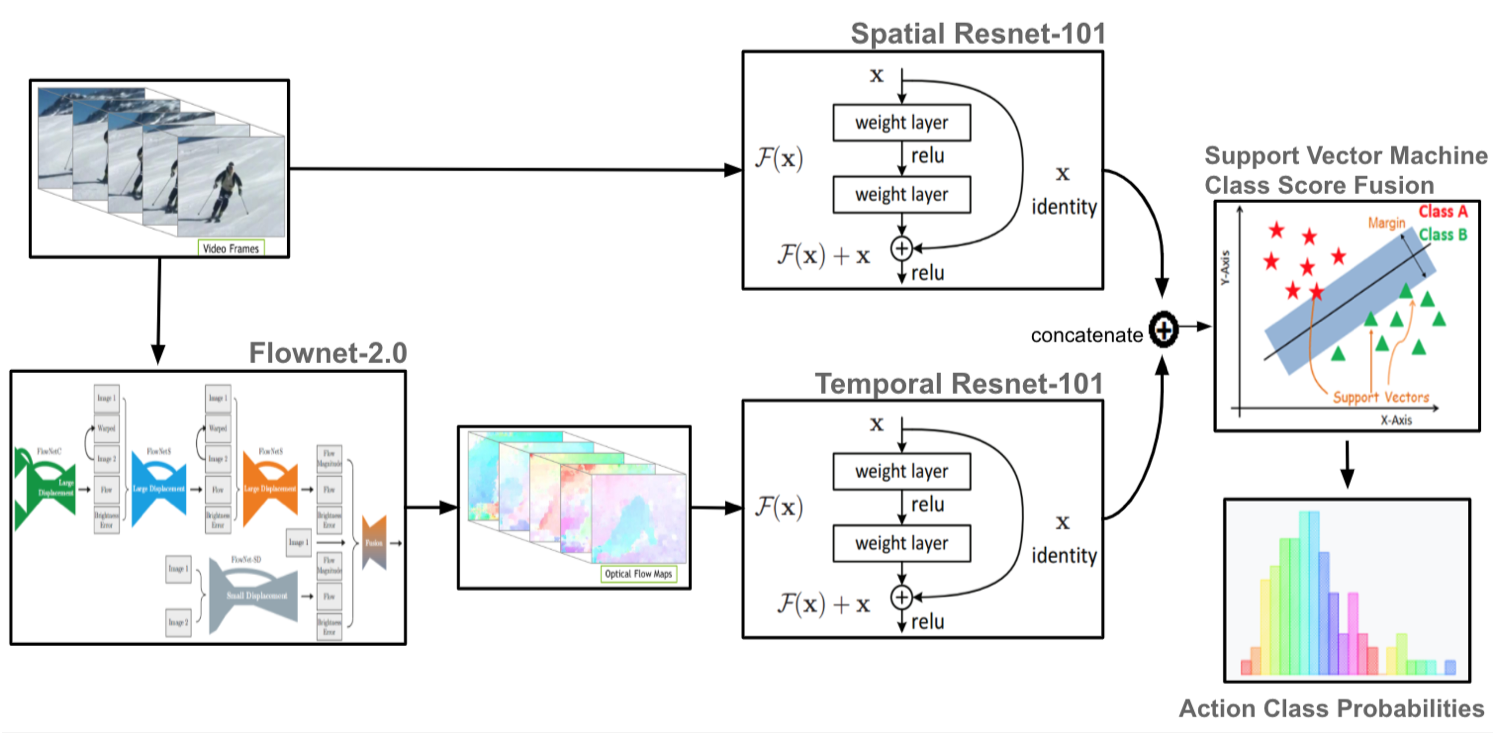}
    \caption{Computer vision network architecture}
    \label{fig:cv_network_arch}
\end{minipage}
\hspace{0.5cm}
\begin{minipage}[b]{0.2\linewidth}
\centering
        \begin{tabular}{@{}lllll@{}}
        \textbf{Architecture} &
        \textbf{FPS} \\
        
        J      & 2 \\
        J+SF   & 5 \\
        C      & 10 \\
        J+SF+R & 11 \\
        C+SF   & 19 \\
        C+SF+R & 45
        \end{tabular}
        \caption{\label{tab:FPS}J=Jetson, SF=Skip Frames, R=Reduced, C=Cloud}
\end{minipage}
\end{figure*}

We propose Low-Light Environment Neural Surveillance (LENS), a proactive, scalable, computer vision based solution for low-light surveillance (Fig. \ref{fig:system_diagram}). LENS implements real-time action recognition algorithms to detect crime in low-light conditions, assists the abilities of civilians to report a crime, and provides potentially expedited response times from government entities in urgent situations. LENS does not rely on the victims or bystanders and is intended to aid victims who are in shock, injured, or unaware of local authorities/blue light systems. We collect and use a dataset tailored for low-light crime, namely, LENS-4 (Section \ref{LENS4}). LENS consists of 3 modules: the computer vision module (Section \ref{ComputerVision}); the IoT module (Section \ref{IoT}); and the user interface module (Section \ref{UserInterface}). The computer vision module performs crime-detection in low light environments via a two-stream approach \cite{DBLP:journals/corr/two_stream}. The IoT module performs post-processing steps to the computer vision module output and relays the post-processed messages to the user interface module. The user interface module receives metadata, such as location of crime, time of crime, and type of crime, and provides a user-to-user notification system. 


\vspace{-.5em}
\section{Previous Work}
\label{sec:previouswork}
\cite{DBLP:journals/corr/two_stream} advanced the action recognition research by training deep Convolutional Neural Networks (CNN) on videos with a two-stream CNN architecture combining the predictions of a spatial and a temporal network. The spatial stream was trained on video frames, while the temporal stream was trained on multi-frame dense optical flow, using UCF-101 and HMDB-51 datasets. Building upon \cite{DBLP:journals/corr/two_stream}, \cite{DBLP:journals/corr/temporal_networks} proposed the Temporal Segment Network (TSN) and improved the two-stream CNN architecture in terms of initialization and training setup. Nevertheless, TSN suffers from the computational bottleneck in calculating the optical flow, preventing real-time application. \cite{DBLP:journals/corr/quo_vadis} proposed the Inflated 3D Convnet (I3D) architecture, a two-stream CNN employing 3D convolutions. 
As 3D convolutions lead to an explosion in the number of network parameters, I3D requires training on large datasets such as Kinetics-400 \cite{DBLP:journals/corr/KayCSZHVVGBNSZ17}. The next steps in action recognition are to leverage spatio-temporal features with 3D residual networks, which has shown promise in recent works such as \cite{DBLP:journals/corr/abs-1708-07632}.

\cite{DBLP:journals/corr/spatio_temporal_conv,DBLP:journals/corr/real_time_action_rec} focused on real-time action recognition. \cite{DBLP:journals/corr/spatio_temporal_conv} developed the C3D architecture, a spatio-temporal neural network using 3-dimensional CNNs, where the temporal information between video frames is implicitly calculated. C3D performs action recognition at a frame rate of 313.9, at the cost of degraded detection performance compared to TSN. \cite{DBLP:journals/corr/real_time_action_rec} circumvented this by calculating optical flow via pre-extracted motion vectors from compressed videos and achieved a frame rate of 390.7.

In only a few recent works, action recognition has been successfully explored in dark environments with infrared and thermal cameras \cite{deeprep,batchuluun2019action}. Nevertheless, using infrared and thermal cameras have several drawbacks and lead to requiring additional technical work compared to standard cameras. \cite{deeprep} require multiple video streams, while \cite{batchuluun2019action} require significant pre-processing of video frames with cycle-consistent generative adversarial networks (CycleGANs) and use a long-short term memory (LSTM) network to capture temporal information. LSTM and CycleGAN-based methods lead to slow inference time and have less possible future improvement, preventing real-time crime detection. Moreover, both \cite{deeprep} and \cite{batchuluun2019action} analyze datasets that have no human-to-human and human-object interaction, making the learning task significantly less complex compared to crime detection. More complex tasks including crime detection bring about the need for normal rather than thermal images, and allow the flexibility of easily extending the application in day-time.
    
We differ from the previous works by proposing an end-to-end action recognition system for accurate real-time crime detection for low-light scenarios. Furthermore, as the state-of-the-art datasets for crime detection are insufficient for low-light environments, we collect and use an action recognition dataset tailored to this application. 

\section{Computer Vision}
\label{ComputerVision}
The LENS computer vision algorithm consists of two parts: dense optical flow calculation and action recognition. We use FlowNet2.0 \cite{DBLP:journals/corr/flownet2} for dense optical flow calculation, and  a two stream approach \cite{DBLP:journals/corr/two_stream} for action recognition, similar to those discussed in Section \ref{sec:previouswork}. The overall network architecture is depicted in Fig. \ref{fig:cv_network_arch}.

\vspace{-1em}
\subsection{Optical Flow}
Optical flow is calculated for the temporal stream of our two-stream approach. We employ FlowNet2.0-CSS, which is shown to operate at high FPS for optical flow calculation without a significant degradation in end-point-error (EPE) compared to FlowNet2.0 \cite{DBLP:journals/corr/flownet2}.

\begin{figure}[!t]
        \centering
        \includegraphics[width=.45\linewidth]{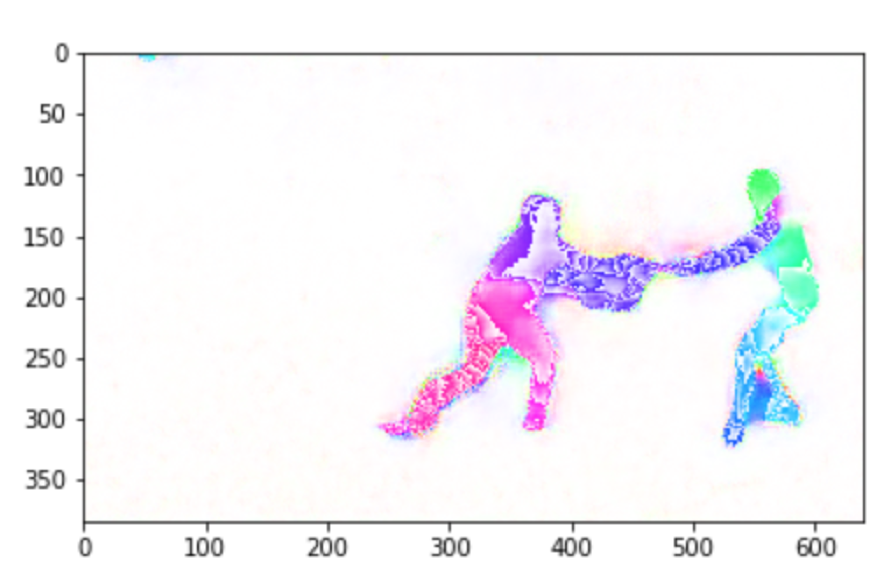}\quad\includegraphics[width=.45\linewidth]{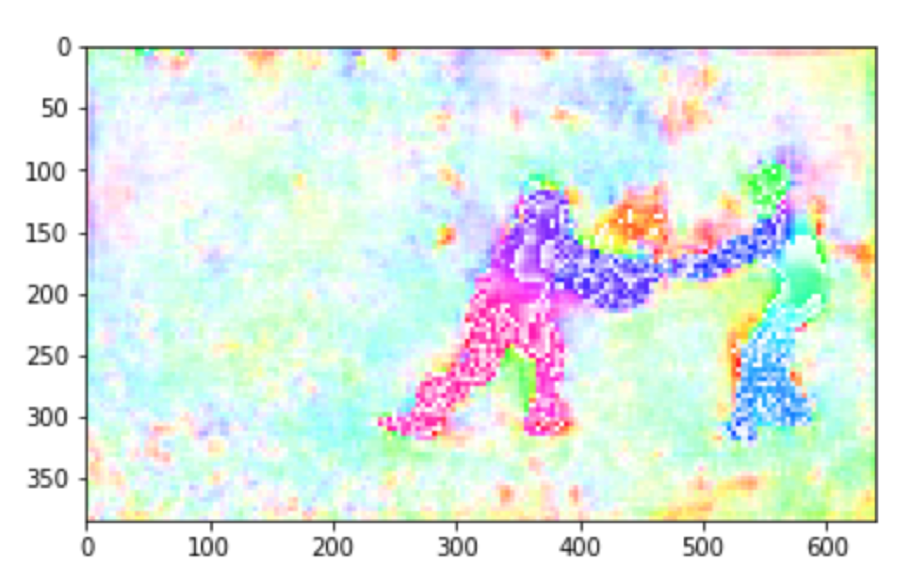}
        \caption{Optical flow output from FlowNet2 (left) and FlowNet2-CSS (right)}
        \label{fig:opf_compare}
        \vspace{-.5em}
    \end{figure}

FlowNet2.0 architecture consists of layered networks, each of which calculates or refines the optical flow. The first three layers calculate large displacement optical flow, which translates to large movements and requires less accuracy, while the last layer calculates small displacements corresponding to fine movements and details. 
The small displacement layer is of relevance as subjects may only be a few pixels, or the movements we are interested in may not be very large. At the output of the network, the last large displacement layer and the single small displacement layer are concatenated to produce the final prediction. Fig. \ref{fig:opf_compare} shows the output of FlowNet2.0 and FlowNet2.0-CSS applied on an example image in the LENS-4 dataset.

\vspace{-1em}
\subsection{Action Recognition}
The action recognition approach follows \cite{DBLP:journals/corr/two_stream,DBLP:journals/corr/temporal_networks}, where two CNN architecture outputs are fused for final action prediction (Fig. \ref{fig:cv_network_arch}): spatial stream trained on video frames, and temporal stream trained on the dense optical flow estimates provided by FlowNet2.0-CSS. We use a two-stream 2D CNN instead of 3D \cite{DBLP:journals/corr/TranBFTP14} to explicitly calculate optical flow in the temporal stream and avoid poor motion estimates. We replace the spatial and temporal stream architectures in \cite{DBLP:journals/corr/two_stream} by ResNet \cite{DBLP:journals/corr/deep_res_learning}, a novel deep neural network architecture that alleviates vanishing/exploding gradients with residual connections. The softmax predictions from spatial and temporal streams are fused via an SVM with a polynomial kernel of order 5.


\section{IoT Capabilities}
\label{IoT}

\subsection{Network Architecture}
The three main network components are edge computers, cloud, and clients. The edge computers use Amazon FreeRTOS and AWS IoT Greengrass to process the images, perform inference using the computer vision module, and communicate with the cloud. The edge computers are WiFi enabled and communicate to a local router that contains the local network of edge computers. In the cloud, AWS IoT Core and Lambda functions process incoming messages and relay them to the clients. The client is a mobile app that requires authentication with the cloud to consume and send messages and requests. These messages are handled with Representational State Transfer (REST) application program interface (API) calls to populate the mobile app and notify law enforcement.

\vspace{-1em}
\subsection{Cloud Services}
In order for the local board hosting the cameras to communicate with the app, a cloud component is used to handle requests. The local board hosting the camera is WiFi enabled in order to scale to multiple cameras in the same network. If the computer vision module detects criminal activity, the corresponding camera ID, GPS location and clip of the incident is sent to the cloud. From there, the message is relayed to law enforcement through the mobile app. 

Most of the computation for the inference model and triggers are handled by the GPU or processor on the local board hosting the camera, allowing the cloud to simply parse the results and relay messages. 
    

\section{User Interface}
\label{UserInterface}

    \begin{figure}[!h]
        \centering
        \includegraphics[width=8cm]{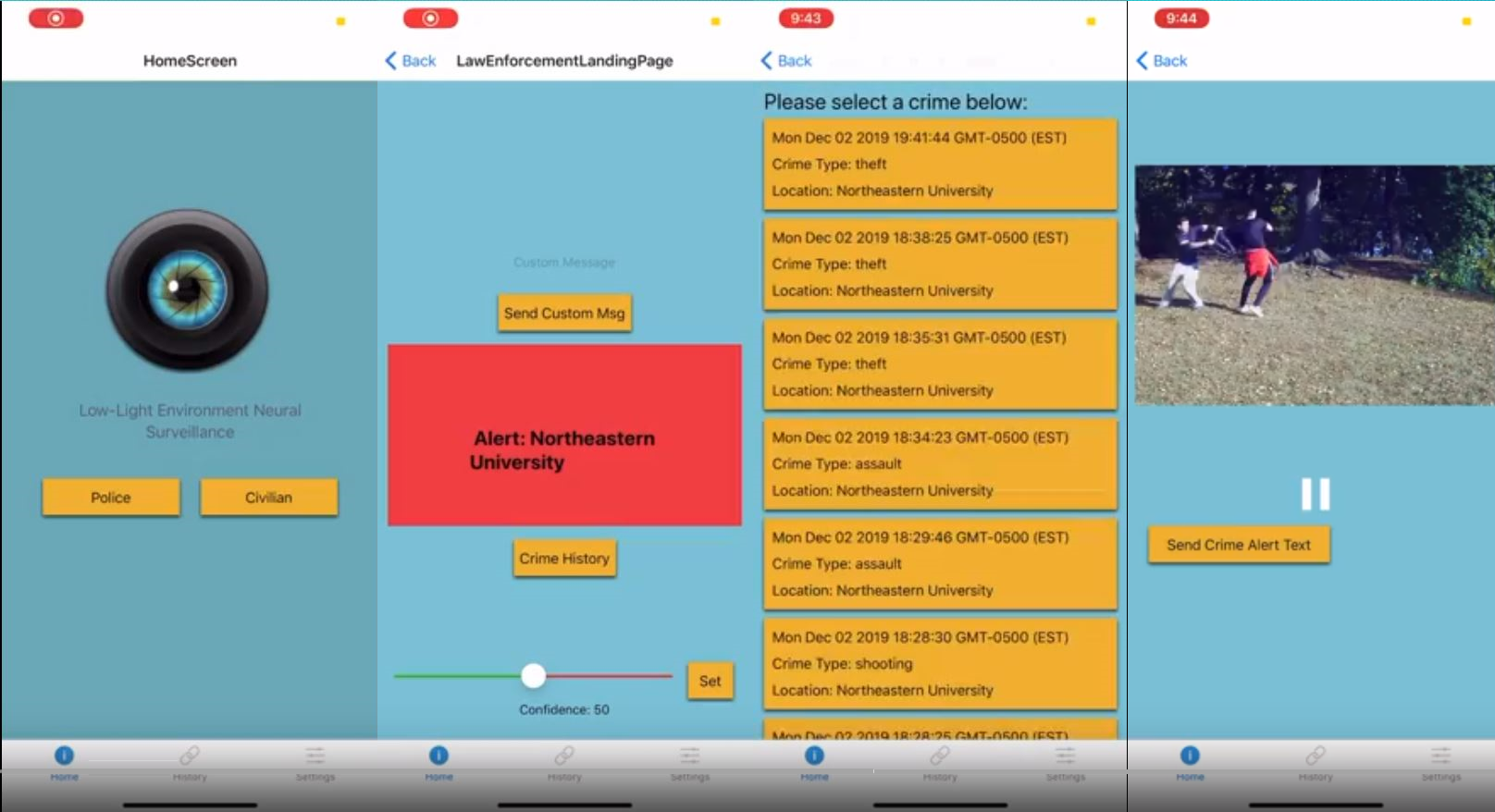}
        \caption{Mobile app mock-ups}
        \label{fig:applayout}
        \vspace{-.5em}
    \end{figure}
    
    \begin{figure*}[!h]
        \centering
        \includegraphics[width=.3\linewidth]{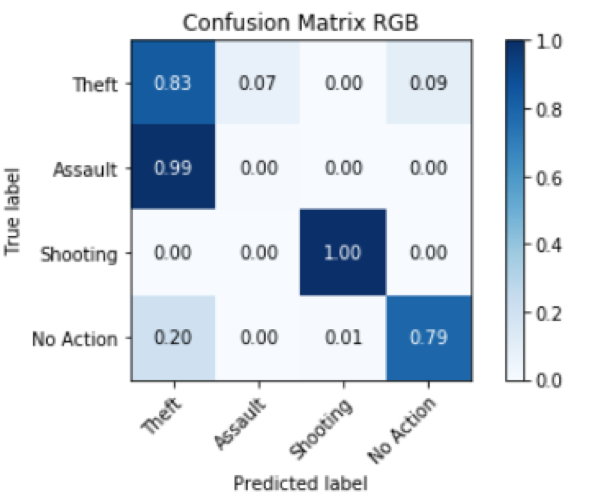}\quad\includegraphics[width=.3\linewidth]{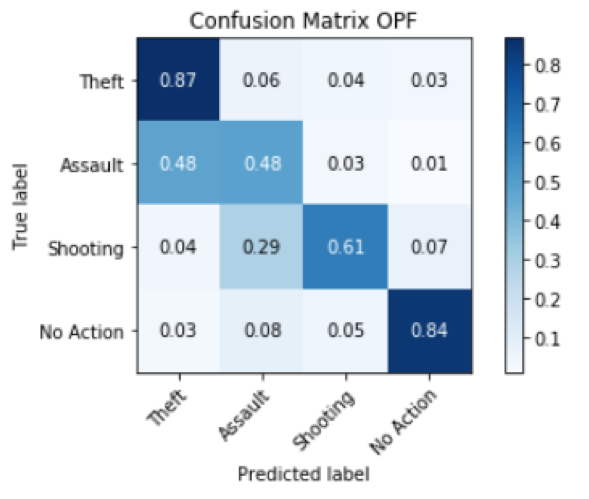}\quad\includegraphics[width=.3\linewidth]{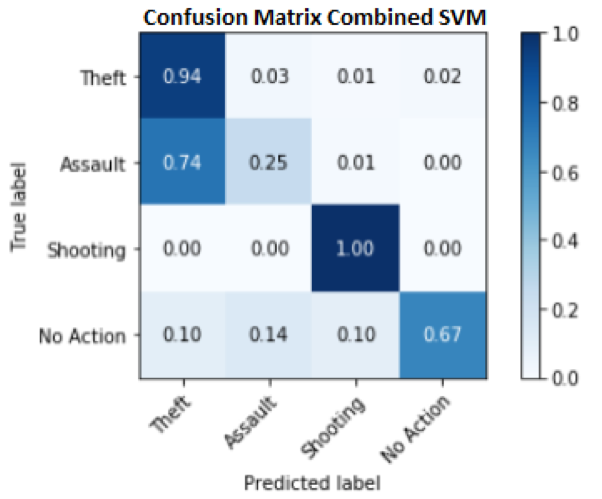}
        \caption{From left to right, confusion matrices for a) spatial stream, b) temporal stream, and c) combined SVM}
        \label{fig:confusion_matrices}
        \vspace{-.5em}
    \end{figure*}
The mobile app (Fig. \ref{fig:applayout}) allows local authorities to monitor and control the LENS system in real-time via a cellular device. There are two levels of user privileges to the mobile app: local authority level and civilian level. The local authority privileges allow features such as requesting video-feed from cameras, and sending out push notifications/text messages to civilians. Civilians still have access to the crime log, but cannot access the associated video feed (Fig. \ref{fig:applayout}). Notifications are pushed to a local authority's device with metadata such as location of crime in relation to location of cameras, time of incident, and type of crime. The local authorities may then request the video feed of the crime that occurred (Fig. \ref{fig:system_diagram}) by navigating to the crime-log in the mobile app (Fig. \ref{fig:applayout}). The local authority may send a push notification/text message to civilians with the civilian-mobile app. 
    
An important feature of the mobile app is the confidence threshold slider (Fig. \ref{fig:applayout}). This slider controls the precision and recall trade-off of crime detection. A high precision and low recall implies that the local authorities will receive a few confident crime detections (low false-positive rate), but not receive as many notifications. A low precision and high recall implies that the local authorities will receive many uncertain crime detections (high false-positive rate), but receive many notifications. This trade-off is left to the local authorities' discretion and allows them to have more control.
    
The app is created using React Native and node.js for easy deployment across iOS, Android, and the web as well as simple integration with AWS.

\vspace{-1em}
\section{Experiments}

\subsection{LENS-4 Dataset}
\label{LENS4}
Although there are numerous publicly available computer vision datasets, less than 2\% of them feature low-light data \cite{DBLP:journals/corr/abs-1805-11227}. On top of this, our application requires crime data captured in low-light. Thus, we create our own dataset, \textit{LENS-4}, for crime detection in low-light. The LENS-4 dataset, explained in detail in table \ref{tab:DatasetSpecs}, has the same structure as the UCF-101 dataset, where the low-light videos are recorded with an ELP Sony IMX322. All clips have a fixed frame rate of 30 FPS and resolution of 640x480 or 1920x1080.

    \begin{table}[!h]
        \centering
        \addtolength{\leftskip}{-1.5cm}
        \addtolength{\rightskip}{-1.5cm}
        \begin{tabular}{@{}ll@{}}
        
        \small Actions & \small 4 \\
       \small Clips & \small 3660 \\
        \small Groups per Action & \small 20\\
        \small Clips per Group & \small 45 \\
        \small Total Duration & \small 244 mins \\
        \small Min Clip Length & \small 3 sec \\
        \small Max Clip Length & \small 4 sec \\
        \small Frame Rate  & \small 30 \\
        \small Resolution  & \small 640*480, 1920*1080 \\
        \small Audio & \small No\\
        
        \end{tabular}
        \caption{\label{tab:DatasetSpecs}LENS-4 Dataset Specs}
    \end{table}
    

The ELP Sony IMX322 has both high SNR and high dynamic range capabilities. The crop CMOS IMX322 is rated to perform in 0.01 lux minimum illumination, which satisfies low-light requirements for collecting data. We preferred the ELP Sony IMX322 over an infrared (IR) camera or a thermal camera for several reasons. Thermal cameras have two disadvantages: a halo effect on objects with high temperature, and temperature similarities between background and objects. Furthermore, IR and thermal cameras are unable to differentiate objects due to requiring heat signature \cite{batchuluun2019action}. Finally, the ELP Sony IMX322 is functional in daytime as well as nighttime, allowing LENS to operate 24/7, while IR and thermal cameras may deliver brighter images in complete darkness, losing pixel-level information and details. 

\vspace{-1em}
\subsection{Training Setup}
Temporal and spatial stream ResNet networks are initialized with weights pre-trained on the ImageNet dataset \cite{deng2009imagenet}. The first convolutional layer weights for the temporal stream is initialized with cross-modality weight initialization \cite{DBLP:journals/corr/temporal_networks}.

We employ data augmentation, batch normalization and dropout on several layers of the CNN architectures as regularization to prevent overfitting. Several data augmentation techniques including random cropping, resizing, channel normalization, and scale/aspect jittering are employed during training and testing.

The spatial CNN parameters are optimized using stochastic gradient descent with momentum set to 0.9 \footnote{Code and LENs-4 dataset are publicly available at https://github.com/mcgridles/LENS}.The initial learning rate is set to 5$x10^{-4}$ and decayed over multiple epochs using a Plateau learning rate scheduler with patience set to 1. For every video in a mini-batch of 64 videos, 3 frames are randomly sampled within equally spaced temporal windows and a consensus of the frame predictions provides a video-level prediction for calculating the loss.

The temporal CNN parameters are also optimized using stochastic gradient descent with 0.9 momentum. The initial learning rate is set to 1$x10^{-2}$ and decayed over epochs using a Plateau learning rate scheduler with patience set to 3. For every mini-batch, 64 videos are randomly selected and 1 stacked optical flow is randomly sampled in each video. 

The temporal and spatial streams are initially trained on the UCF-101 dataset, with pre-computed TVL1 optical flow. We improve the two-stream action recognition pipeline accuracy on split 1 of UCF-101 compared to \cite{DBLP:journals/corr/two_stream} (table \ref{tab:IndividualStreamResults}). Finally, the spatial and temporal streams are fine-tuned using transfer learning on the LENS-4 dataset. 

    \begin{table}[!h]
        \centering
        \addtolength{\leftskip}{-3.5cm}
        \addtolength{\rightskip}{-3.5cm}
        \begin{tabular}{@{}lllll@{}}
        \textbf{} & 
        \textbf{Spatial Acc.} & 
        \textbf{Temporal Acc.} &
        \textbf{Two Stream Acc.} & \\
        
        \cite{DBLP:journals/corr/two_stream} & 72.7 & 82.1 & 86.2 \\
        Ours                                 & 82.1 & 82.3 & 88.8 \\
        \end{tabular}
        \caption{\label{tab:IndividualStreamResults}Individual Stream and Combined Split 1 Accuracy}
    \end{table}
    
The SVM is trained using 5-fold cross validation, and hyper-parameters are selected by randomized grid search.

All models are trained on Google Cloud, as large memory and computation capabilities are required to train deep neural networks. A virtual machine is deployed using the Deep Learning VM image in the Google Cloud Marketplace on the Google Cloud Compute Engine, where the PyTorch, FastAI, and NVIDIA configuration settings are applied. Due to the training datasets taking over 100 Gigabytes (GB) in memory,  a 500 GB zonal persistent disk is added and mounted to the VMs for dataset storage. The GPU used is an NVIDIA Tesla P100.

\vspace{-1em}
\subsection{Action Recognition}
For the spatial stream, we achieve 65.5\% accuracy after fine-tuning on the LENS-4 dataset. A similar process for the temporal stream achieves 70.3\% accuracy, confirming that motion plays an inherent role for action recognition. The last step in the inference pipeline, the SVM, combines the predictions from each of the two streams for final crime inference, where the nonlinear predictor improves the accuracy of crime detection up to 71.5\% (table \ref{tab:Accuracies}).

    \begin{table}[!h]
    \centering
    \addtolength{\leftskip}{-3.5cm}
    \addtolength{\rightskip}{-3.5cm}
    \begin{tabular}{@{}lllll@{}}
    
    \textbf{} & 
    \textbf{Train Acc.} & 
    \textbf{Val. Acc.} &
    \textbf{Test Acc.} & \\
    
    Spatial Stream    & 98 & 71 & 65.5  \\
    Temporal Stream   & 90 & 87.1 & 70  \\
    Two Stream + SVM               & - & - & 71.5   \\
    
    \end{tabular}
    \caption{\label{tab:Accuracies}LENS-4 Accuracy}
    \end{table}

\begin{figure}[!h]
        \centering
        \includegraphics[width=8cm]{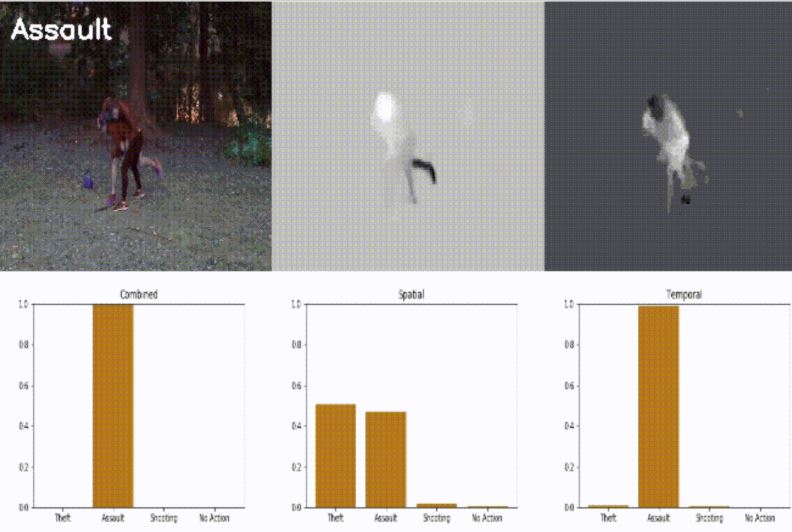}
        \caption{Clockwise from top left: RGB frames, optical flow (horizontal), optical flow (vertical), spatial predictions, temporal predictions, and combined predictions}
        \label{fig:AR_arch}
    \end{figure}

An example of LENS crime detection (at dusk for assault) inference is shown in Fig. \ref{fig:AR_arch}.

The RGB confusion matrix (Fig. \ref{fig:confusion_matrices}a) indicates that the spatial stream cannot distinguish between assault and theft. Assault and theft are hard to distinguish because they are not mutually exclusive classes, but rather, assault may be seen as a subclass of theft. Even state-of-the-art computer vision tasks such as object detection \cite{DBLP:journals/corr/RedmonF16} are reported to have a difficulty in distinguishing non-mutually exclusive classes. The temporal stream resolves the spatial stream difficulties of distinguishing between assault and theft, as the motion that occurs for the two actions is typically different (as when a thief tries to steal a bag, both actors grab the bag and pull in a seesaw manner). However, the temporal stream performs worse than the spatial stream on shooting detection, as the OPF confusion matrix indicates that the temporal network confuses shooting with assault (Fig. \ref{fig:confusion_matrices}b). Again, shooting and assault are not mutually exclusive classes, and the motion associated with shooting shares similarities with the motion of assault, and the temporal stream only learns from motion, i.e. the temporal stream does not learn about what the gun looks like, only the motion associated with movements of an actor holding a gun. The SVM confusion matrix (Fig. \ref{fig:confusion_matrices}c) shows that the deficiencies from each individual stream are compensated by the SVM fusion (table \ref{tab:percentchanges})
.
    \begin{table}[!h]
        \centering
        \addtolength{\leftskip}{-3.5cm}
        \addtolength{\rightskip}{-3.5cm}
        \begin{tabular}{@{}lllll@{}}
        
        \textbf{} &
        \textbf{Theft} & 
        \textbf{Assault} & 
        \textbf{Shooting} &
        \textbf{No Action} \\
        Spatial  & 8 & -48 & 64 & -20 \\
        Temporal & 13 & $\infty$ & 0 & -15 \\
    
        \end{tabular}
        \caption{\label{tab:percentchanges}Percent Changes of spatial stream and temporal stream to SVM fusion}
    \end{table}

Overall, we achieve an accuracy of 71.5\% for crime detection in near pitch black environments at a frame rate of 19 FPS (section \ref{framerate}). \emph{This is a breakthrough, as crime detection in low-light environments with real-time inference capability has not yet been explored}.

\vspace{-1em}
\subsection{Frame Rate}\label{framerate}
For total inference time, we achieve approximately 10 FPS running on the NVIDIA Tesla P100 in the cloud. Skipping frames improves the FPS without loss of accuracy, since the motion captured between two sequential frames when recording at 30 FPS is significantly redundant. As many as three or four frames could be skipped before performance started to suffer, which would amount to a 3-4x increase in speed. On the NVIDIA Jetson TX2, we are able to achieve approximately 5 FPS due to the decreased processing power. However, we improve this in a few ways. The first is by skipping frames as mentioned previously. We also manage to stream from the Jetson to the cloud and perform the inference on the cloud. While this does not run as fast as native video on the cloud, it still improves the inference time and would be worth developing further. By dropping every other frame we achieved 20 FPS (Table \ref{tab:FPS}).

%
%
%
    

\section{Conclusions and Future Work}
We introduce LENS, a complete and modular system that serves as a tool for law enforcement to proactively combat criminal activity in low-light areas. It has been shown that LENS runs in real-time, operate in a low-light environment on a single camera (unlike many current architectures), and is scalable with the cloud infrastructure. LENS allows police to be immediately notified of potential criminal activity detected, and does so with a high degree of accuracy.

Future improvements to this project to create a more robust, faster, and accurate system include: 1. collecting a larger and more diverse dataset than LENS-4,  2. using Active Learning in the LENS-human feedback loop, and 3. developing shallower LENS architecture.

The size and diversity of actors in the LENS-4 is extremely small, which could lead to over-fitting models if regularization techniques are not carefully incorporated. Actors and cameramen should be hired to record a large volume of clips, in varying scenery, with actors performing fake shootings, thefts, and assaults. 

Active learning is a study of machine learning where the algorithm may interactively query the user to obtain a user-label output for new data points. Active learning allows local authorities to receive clips of the crime that occurs, and if it is a false positive/incorrect crime label the user could correctly label the video clip and then LENS could be retrained in an online manner with the new information. 

Finally, we will look at developing a shallower LENS architecture for faster inference time without reducing the accuracy of the model. 











\vspace{-1em}
\begin{small}
\bibliography{refs.bib}
\bibliographystyle{IEEEbib}
\end{small}

\end{document}